\documentclass[conference,harvard,brazil,english]{sbatex}
\usepackage[utf8]{inputenc}
\usepackage{ae}
\usepackage{graphicx}
\usepackage{amsmath}
\usepackage{comment}
\usepackage[T1]{fontenc}
\usepackage{url}
\usepackage{booktabs}
\usepackage{color}
\usepackage{amssymb}
\usepackage{amsmath}
\usepackage{float}
\usepackage{enumitem}
\usepackage{color}
\begin{document}
%
%
\title{Robotic Tankette for Intelligent BioEnergy Agriculture: Design, Development and Field Tests$^{1}$}
%
%

\author{Marco F. S. Xaud}{mafernan@nmbu.no}
\address{Norwegian
University of Life Sciences, \\ Faculty of Science and Technology, \\ P.O. Box 5003, NO-1432,  \AA s, Norway.}
\author{Antonio C. Leite}{antonio@ele.puc-rio.br}
\address{Pontifical Catholic University
of Rio de Janeiro, \\ Department of Electrical Engineering, \\ Postal code 22451-900 Rio de Janeiro RJ, Brazil.}
\author[1]{Evelyn S. Barbosa$^{2}$}{evelynsoaresb@poli.ufrj.br}
\author[1]{Henrique D. Faria$^{2}$}{hd.faria@poli.ufrj.br}
\author[1]{Gabriel S. M. Loureiro$^{2}$}{gloureiro@poli.ufrj.br}
%
%
%
\author[1]{P\aa l J. From}{pal.johan.from@nmbu.no}
%

\twocolumn[

\maketitle

\selectlanguage{english}
\begin{abstract}
In recent years, the use of robots in    agriculture
has been increasing mainly due to the high demand of productivity, precision and efficiency, which follow the  climate change effects and world population growth. 
Unlike conventional agriculture,
sugarcane farms are usually regions with dense vegetation, gigantic areas, and subjected to extreme       weather
conditions, such as intense heat, moisture and rain. 
TIBA - Tankette for Intelligent BioEnergy Agriculture -    is
the first result of an R\&D project which strives to  develop
an autonomous mobile robotic system for carrying out        a
number of agricultural tasks in sugarcane fields. 
The proposed concept consists of a semi-autonomous,
low-cost, dust and waterproof tankette-type vehicle,  capable
of infiltrating dense vegetation in plantation
tunnels and carry several sensing systems, in order        to
perform mapping of hard-to-access areas and collecting samples. 
This paper presents an overview of the robot       mechanical
design, the embedded electronics and
software architecture, and the construction of a        first
prototype. 
Preliminary results obtained in field tests validate      the
proposed conceptual design and bring about several challenges
and potential applications for robot autonomous navigation, as
well as to build a new prototype with              additional
functionality. 
\end{abstract}
\selectlanguage{brazil}
\begin{abstract}
Nos últimos anos, o uso de sistemas robóticos na  agricultura
tem aumentado principalmente devido à alta demanda em  termos
de produtividade, precisão e eficiência, causadas       pelos
efeitos das mudanças climáticas e pelo crescimento         da
população mundial.
Ao contrário da agricultura convencional, fazendas de cana de açucar são geralmente  regiões com
densa vegetação, áreas gigantescas, e sujeitas a condições meteorológicas extremas,    como
calor, umidade e chuva intensa.
TIBA - \emph{Tankette for Intelligent BioEnergy Agriculture} -
é o resultado preliminar de um projeto de P\&D cujo objetivo é desenvolver um sistema robótico móvel e   autônomo
capaz de realizar diversas tarefas agrícolas em fazendas   de
cana-de-açúcar.    
O conceito proposto consiste de um pequeno veículo       tipo
tanque, de baixo custo, à prova de poeira e água, capaz de se
infiltrar em densas vegetações em túneis de plantação, carregando diversos sensores e dispositivos   para
realizar tarefas de mapeamento em áreas de difícil acesso   e
coleta de amostras. 
Este trabalho apresenta uma visão geral do projeto   mecânico
do robô, uma descrição da arquitetura de software           e
eletrônica embarcada e a construção de um primeiro protótipo.
Resultados preliminares obtidos em testes de campo  validam o
projeto conceitual proposto e levantam vários desafios      e
potenciais aplicações para navegação autônoma do robô     bem
como para construir um novo protótipo com     funcionalidades
adicionais.
\end{abstract}

]


\selectlanguage{english}

\footnotetext[1]{This work was partially supported by the UTFORSK Partnership Programme from The Norwegian Centre for International Cooperation in Education (SIU), 
project number UTF-2016-long-term/10097.}
\footnotetext[2]{Visiting Research Student at the Norwegian University of Life Sciences.}
\section{Introduction} 
In recent years, the use of robots in agriculture          or 
``agbots'' has been growing significantly due to the     high
demand for increased productivity and precision, which    was
brought about by climate changes and their effects on     the
environment \cite{Billingsley_2008}. 
Moreover, the growth of the world's population has  motivated
farmers to seek an increase in the efficiency of         food
production, for instance, reducing the waste of inputs    and
increasing the productivity of small cultivated areas, at the
lowest possible cost \cite{Edan_2009}.
Currently, there are more than 50 different types   of mobile
robots in the world deployed for agricultural purposes.
These robots, in general, have specialized accessories, tools
and arms to perform a wide range of agricultural tasks
\cite{Bechar_2017,Grimstad2017}.
This number gives us an idea of how the potential of         robotic
technology is embedded in agricultural systems.
Following this trend, it is possible to find          devices
vacuuming apples off the trees in the US, octopus-like robots for
collecting strawberries in Spain, and machines feeding        and
milking cows in the UK. These are just a few examples of      how
robots are taking over fields around the world. 
These concerns about food production do not            belong
exclusively to Europe and can also be considered in      many
countries of Central and South America, particularly,      in
Brazil. 
Although, the Brazilian agricultural industry has a      high
degree of automation for the planting and harvesting       of
grains and sugarcane in large areas, farmers still do     not
use autonomous robotic systems to perform basic and   complex
agricultural tasks in small areas, such as, vegetable gardens
and orchards. 
Some basic agricultural tasks include sowing,    fertilizing,
and irrigation, while some complex agricultural tasks consist
of harvesting fruits, killing weeds and plant     phenotyping
\cite{Bac_2014,Midtiby_2016,Li_2014}. 

Sugarcane has emerged as an important alternative to fossil fuels since, besides to produce sugar, it can  be
used as a clean, renewable source of energy reducing      the 
petroleum use - and consequently greenhouse gas   emissions - 
and also to generate other useful products such           as, 
bioelectricity and bioplastics \cite{Davis2009}. 
Sugarcane farms usually need gigantic areas for     planting,
which requires many employees and machinery, as well as   high
costs for maintenance and logistics. Moreover, most        of
sugarcane tasks are laborious and tedious, susceptible     to
human errors and performed in a harsh environment subject  to
bad weather conditions, dense vegetation, and wild life.
Some of the relevant tasks include: cataloging of plants,
identification of weeds/pests and misplanting  (empty spots),
classification of soil types, and evaluation of plant health.
\begin{figure}[!htpb] 
	\centering
	\includegraphics[angle=0,width=1.0\columnwidth]{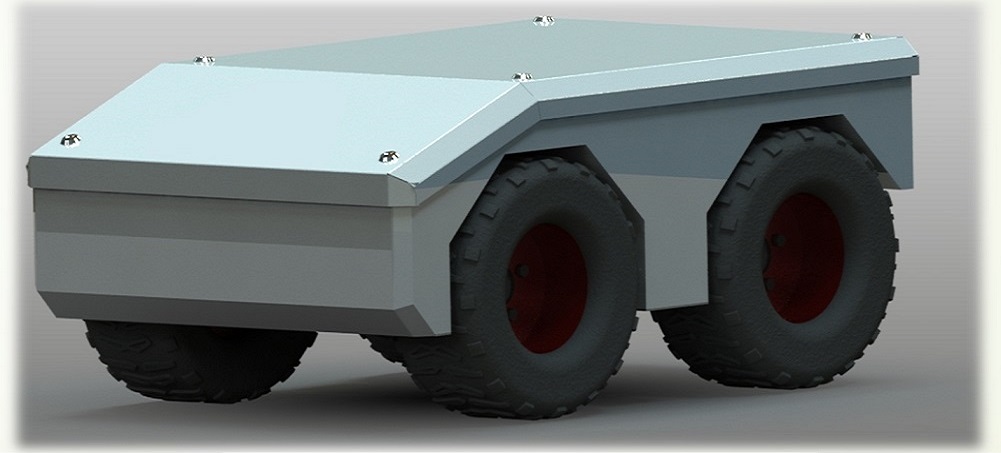}
	\caption{Conceptual design of the TIBA robot.} 
	\label{FIG:MEC1}
\end{figure} 

Following this motivation, in this proposal we aim        to
develop the conceptual design and build
the prototype of a mobile robot to 
perform autonomous tasks of mapping difficult-to-access areas     and
collecting soil samples in sugarcane farms.
TIBA - Tankette for Intelligent BioEnergy Agriculture - 
is an alternative solution to deploying several robot
units in a swarm-based approach to cover a wide area, reducing   or
eliminating the employment of man-power under       unhealthy
environments and enhancing the farm logistics. To the best our knowledge, TIBA (Fig.\,\ref{FIG:MEC1}) is the first prototype of   a
4-wheeled mobile robot to be developed to carry out precision agricultural tasks on
sugarcane plantations. Similar systems have been tested in vineyards, orchards and corn production in the US (e.g., Rowbot).   
%

\section{Robot for Sugarcane Fields}
%
In this section, we describe
%
the main aspects of the        proposed
conceptual design of the TIBA robot in terms of    mechanical
structure, locomotion system, embedded electronics,       and
software architecture. 
This design was conceived based on the main challenges  found
in sugarcane fields, such as tight and dense 
plant rows,
soil mostly composed of sand ($66.7\%$)                   and 
clay/mud ($33.3\%$), and possibility of plant leaves  getting
trapped in the robot parts. 
%
\subsection{Mechanical Design} %
The mechanical structure was designed to provide a       very
compact solution yet capable to yield 3    degrees-of-freedom,
translation $x,y$ and the rotation $\theta$. The robot is a     4-wheeled
small tank composed of a $3\,mm$ thick steel chassis      and 
$1.5\,mm$ aluminum walls (Fig.\,\ref{FIG:MEC1}). There is also
an aluminum lid on the top, in order to cover the robot
interior, protect it against water and dust - which is   also
provided by rubber tapes along the lid perimeter -        and
accommodate sensors or other devices over its surface,   such
as laser scanners, solar panels, etc. The robot carcass   has
also customized holes to provide mechanical accommodation and
view of sensors and cameras to the outside. 
The robot interior accommodates the mechanical  arrangement---composed of two motors, two gearboxes,          belts/chains,
pulleys/sprockets, wheel bearings and adapters---and      the
support for the electronics system, which is composed      of
aluminum profiles in such an arrangement that does        not 
interfere with the mechanical components. 
%

\begin{figure}[htpb] 
	\centering
	\includegraphics[angle=0,width=1.0\columnwidth]{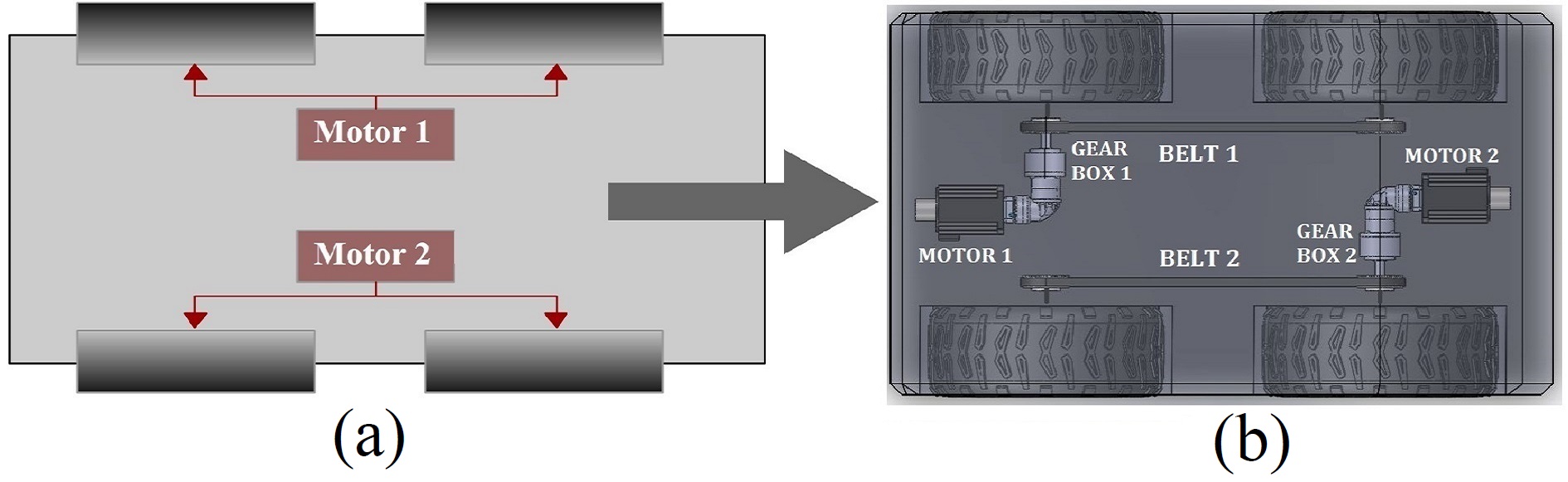}
	\caption{Driving configuration - (a) skid-steering; (b) TIBA mechanical arrangement.  } 
	\label{FIG:SKIDSTEERING}
\end{figure}

The main motivation for this project was the reduction     of
costs. Thus, a driving configuration that provides the   best
mobility with the lowest number of motors and           other
mechanical parts had to be chosen. The          skid-steering
configuration, as shown in
Fig.\,\ref{FIG:SKIDSTEERING}(a), 
%
was selected  as
a suitable and low-cost solution, which only demands      the
employment of two motors, being the curved movements achieved
by applying differential speeds in each side. In    addition,
the trajectory control of skid-steering robots is     already established in the literature, and          its
implementation can be based on several recent works, such  as 
\cite{control1}. In TIBA, 
one motor is designated to drive the two wheels of the    same
side, as shown in 
Fig.\,\ref{FIG:SKIDSTEERING}(b). 
%
Those wheels         are
connected by a belt, which provides the same rotation rate to
them. Finally, one gearbox connect to each motor provides the
torque increase and speed reduction. A right-angle    gearbox
model was employed as a solution for         the
confined space.  

Aiming to sustain the cost reduction strategy,            the
manufacturer and models of the motors, gearboxes and   wheels
were selected from the set employed in the existing     robot
Thorvald II \cite{Grimstad2017}, so as to get out of    their
already verified advantages, which include       reliability,
low-cost, robustness, easiness for prototyping,           and
compatibility with other components. 
The wheels are a customized set of tire and rim suitable  for
snow, sand and clay terrain. To define the suitable set    of
motors and gearboxes, a calculation of the needed torque  had
to be performed. Given the selected wheel model -        with
$r_{w}\!=\!20\,cm$ tire radius - and the required operational
terrain - clay and sand - we could estimate the     necessary
linear force to overcome the static friction between the tire
rubber and the ground.

By considering an estimated robot weight of $M\approx130\,kg$, 
the Coulomb friction coefficient and the rolling   resistance
coefficient between an off-road tire and sand              of 
$\mu_{s}\!=\!0.6$ and $C_{rrs}\!=\!0.2$ respectively, and the
same coefficients for wet earth road or clay               of 
$\mu_{c}\!=\!0.55$ and                     $C_{rrc}\!=\!0.08$ 
\cite{wong2008theory}, the necessary force to overcome    the
friction applied in each wheel is given by $F_{w}= N_{w}\,\mu_s$, where $N_{w}$ is the normal force on each wheel, and  $\mu_s$
is the taken as the worst case among the              terrain
possibilities. By considering $g\approx 9.8m/s^2$, the normal
force is approximately given by 
$N_{w}\!=\!Mg/4=130\times 9.8/4=318.5\,N$, 
and hence, $F_{w}\!=\!318.5\times 0.6\!=\!191.1\,N$.      The 
necessary torque $\tau_{w}$ on each wheel is given by $\tau_{w}=F_{w}\,r_{w}$, which yields $\tau_{w}\!=\!191.1\times 0.2\!=\!38.22\,N.m$. 
Finally since each motor drives two wheels at the same   time
and considering that this torque is equally       distributed
between them along the belt, the necessary torque to    drive
each wheel/belt set is given by 
$\bar{\tau}_{b}=76.44\,N.m$.  

Among the available motor options from the selected
manufacturer, the model 3MeN\textsuperscript{\textregistered}
BL840 provides the higher nominal torque                   of 
$\tau_{m}\!=\!16\,kgf.cm\approx 1.57\,N.m$. 
Given that, among the available gearbox options from      the
selected manufacturer, the                              model
APEX\textsuperscript{\textregistered} AER070-050 provides   a
reduction ratio of 50:1. This drives the wheel/belt set  with
an increased torque of $\tau_{b}\!=\!50\times \tau_{m}=78.50\,N.m$, which is more than   the
needed value $\bar{\tau}_{b}$, as calculated previously. 

Without any load, this motor/gearbox combination provides  an
output angular speed of $3000\,rpm$ in the input of      each
gearbox set, and an output angular speed of $3000/50=60\,rpm$
in each wheel, which provides a total linear speed         of
$v\!=\!2\pi\times 60\times 0.2/60\approx 1.26\,m/s$      or 
$v\approx 4.52\,km/h$. Considering the given environment  and
operational requirements for the robot, this is a fairly fast
and safe maximum speed.
%
\subsection{Embedded Electronics Design} 
\begin{figure}[htpb] 
	\centering
	\includegraphics[angle=0,width=1.0\columnwidth]{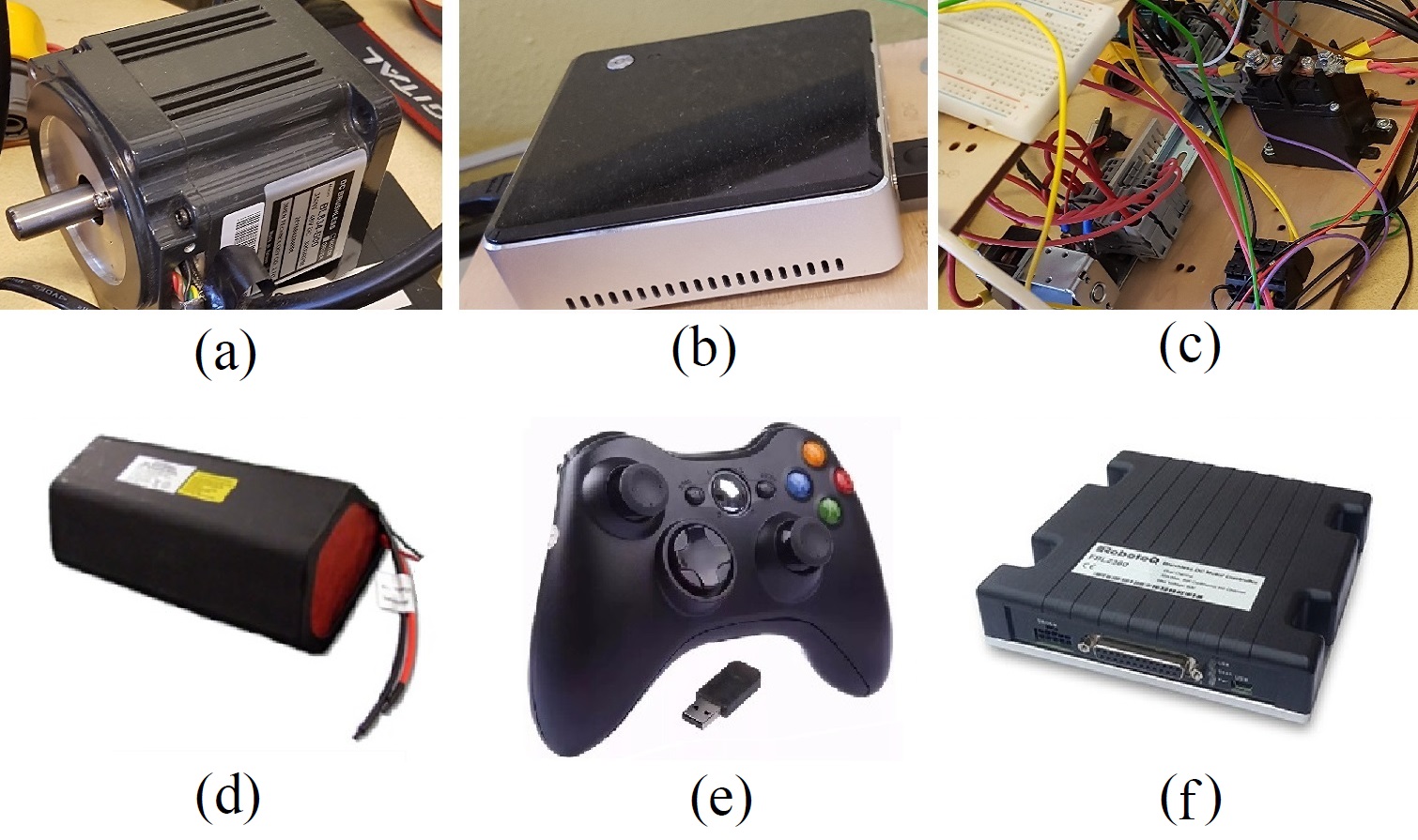}
	\caption{Embedded electronics parts - (a) Motor; (b) NUC computer; (c) DIN rail arrangements; (d) battery pack; (e) joystick; (f) motor driver.}  
	\label{FIG:ELEC1}
\end{figure}

The embedded electronics architecture
of the robot was also designed so as to
comply with the cost and environment requirements, and   thus
several electronic parts from the robot Thorvald II were used,
such as the motor, power driver, mechanical relays, computer,
rugged cable/connectors, and DIN rail (Fig.\,\ref{FIG:ELEC1}).
%
%

The power is supplied by one 48VDC NiMH Battery Pack,     and
distributed into two power buses of 48VDC and 12VDC.      The
former feeds the power controller driver, and the      latter
supplies the computer, peripheral sensors, and              a 
\emph{Vehicle Support System} (VSS) - which is represented in
the first version by an Arduino board. The           computer 
Intel\textsuperscript{\textregistered}                    NUC
NUC6I5SYK Skylake gives a compact, lightweight and  practical
solution for a first prototype, and counts with a      fairly
powerful processor Intel Core i5 and several USB ports.   The
driver Roboteq\textsuperscript{\textregistered} FBL2360    is
also a compact solution with two channels, which allow    the
control of two different motors in the same device. 
The control signals are communicated from the PC to       the
driver via \emph{Controller Area Network} (CAN), which is  an
established and robust network for sensors and      actuators
communication with strict error detection. A set of     DC-DC
converters, low-power relays   and a Gigabit Ethernet switch
gives the functionality of adding/removing any     peripheral
device or sensor from the system, communicate them to     the
computer, and turn them on/off individually to save   energy.

The VSS has the role of this relay activation,        reading
voltage and                                           current
measurements, and communicate them to the PC via CAN. The VSS
is also integrated with sensors for the electronics    system
self-monitoring, such as the internal temperature         and
humidity. The system also counts a touchscreen     panel
mounted on back of the robot, and an XBox wireless joystick   for
the control in manual mode. The panel and the        joystick
antenna are both connected to the PC directly via USB,  being
the cables routed outside the robot through a      waterproof
connector.
All the electronic devices were mounted over a robust    and
firm aluminum structure. The DIN-rail was a robust   solution
for wiring and keeps the connections firm under vibration and 
impact conditions.
\begin{figure}[htpb] 
	\centering
	\includegraphics[angle=0,width=0.95\columnwidth]{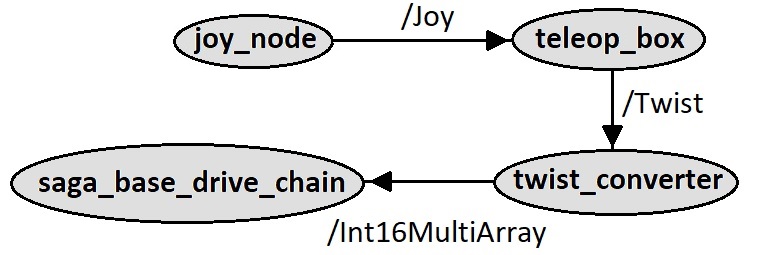}\\
	\caption{Diagram of ROS nodes. } 
	\label{FIG:SOFT}
\end{figure} 
\subsection{Software Architecture}
The robot software is developed on the modular multi-threaded
control framework ROS \cite{quigley2009ros}, which presents a
large community with continuous expansion of software modules (Fig.\,\ref{FIG:SOFT}). 
%
%
The prototype is teleoperated using a Xbox One       wireless
joystick. The teleop\textunderscore box node is   responsible
to map the joystick axis and buttons to the           desired
velocities and gains. 
The velocities can be modeled as an increment of   velocities
on the $xy$ axis. Then, a twist message is publish on a topic.
The twist converter node subscribed to the Twist topic    and
calculates the  real velocities values. Then, it    publishes
them on a topic that                                      the
saga\textunderscore base\textunderscore  drive\textunderscore
chain node is subscribed. Finally, the latter is  responsible
to communicate with the drivers via the CANopen protocol.
%
%
%
\begin{figure}[htpb] 
	\centering
	\includegraphics[angle=0,width=0.95\columnwidth]{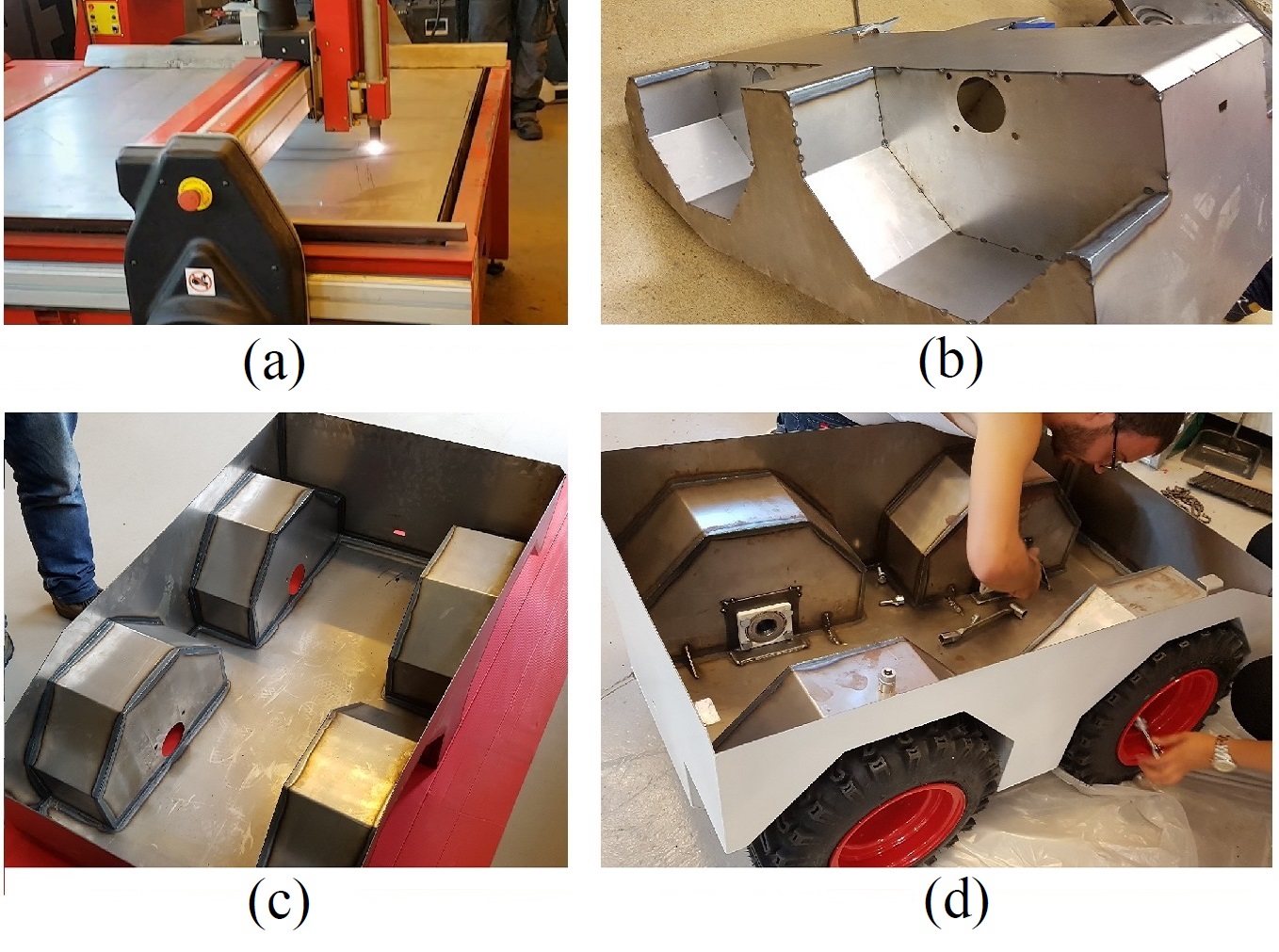}
	\caption{Assembly of early prototype - (a) cutting of steel and aluminum plates by using plasma cutter; (b) welded parts; (c) chassis and carcass; (d) assembly of mechanics system. } 
	\label{FIG:ASSEMBLY1}
\end{figure} 
%
\subsection{First Robot Prototype} 
\begin{figure}[htpb]
	\centering
	\includegraphics[angle=0,width=0.95\columnwidth]{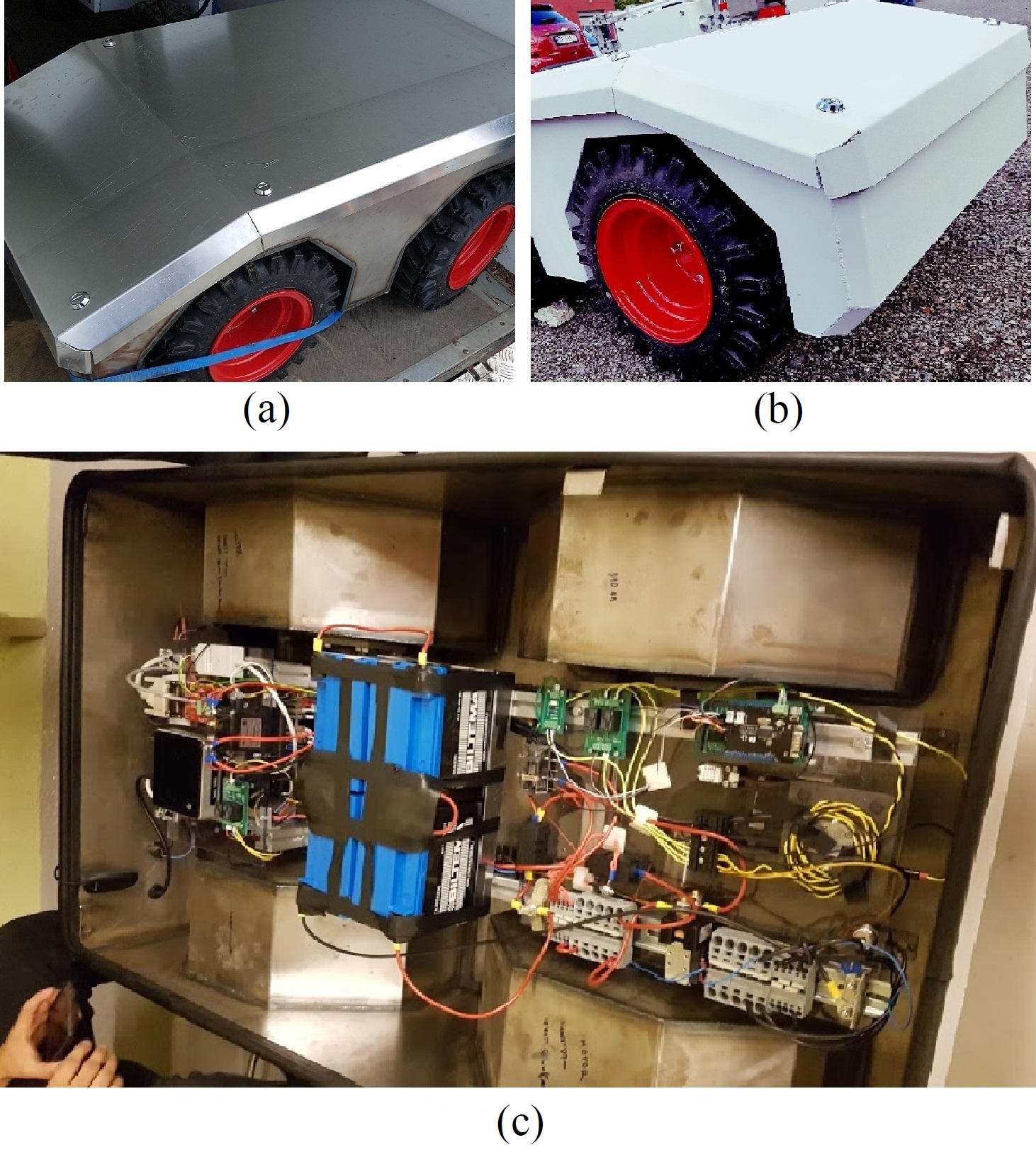}
	\caption{Early prototype - (a) top lid and robot closed; (b) painted prototype; (c) compact electronics system on the support. } 
	\label{FIG:ASSEMBLY2}
\end{figure}

The first prototype, shown in Figures \ref{FIG:ASSEMBLY1} and
\ref{FIG:ASSEMBLY2}, was constructed in order to    validate
all the proposed mechanical concepts, feasibility of      the
implemented skid-steering configuration, robot behaviour   in
the expected terrains, and robustness of the         embedded
electronics system to vibration and environmental conditions. 
The robot chassis and carcass were basically a        welded
assembly of cut and bent metal sheets. Only for this    first
version, some modifications and simplifications          were
implemented due to cost reductions, delays in delivery     of
equipment and assembly pieces, and fulfillment of schedules.

Among the most significant simplifications, the         robot
mechanical structure, except the top lid, was made   entirely
of steel, so that the parts welding could be performed faster.
The desired wheel bearing model could not be found on time
and, thus, an alternative one from Volvo was used, which was
however heavier, and needed several mechanical adapters for
fixation. 
The belts and pulleys were replaced by chains and  sprockets,
since the belts would demand a very large width for the given
torque/power and, hence, the redesign of the      electronics
support structure. 
Finally, due to the lack of available encoders, the     motor
control and odometry were implemented by only using       the
motors hall-effect sensors. 
This condition restricted and hindered the motor     position
control and high accuracy for an odometry model, but it   was
satisfactory for the first tests.    
%
\section{Field Tests} 
\begin{figure}[htpb] 
	\centering
	\includegraphics[angle=0,width=1\columnwidth]{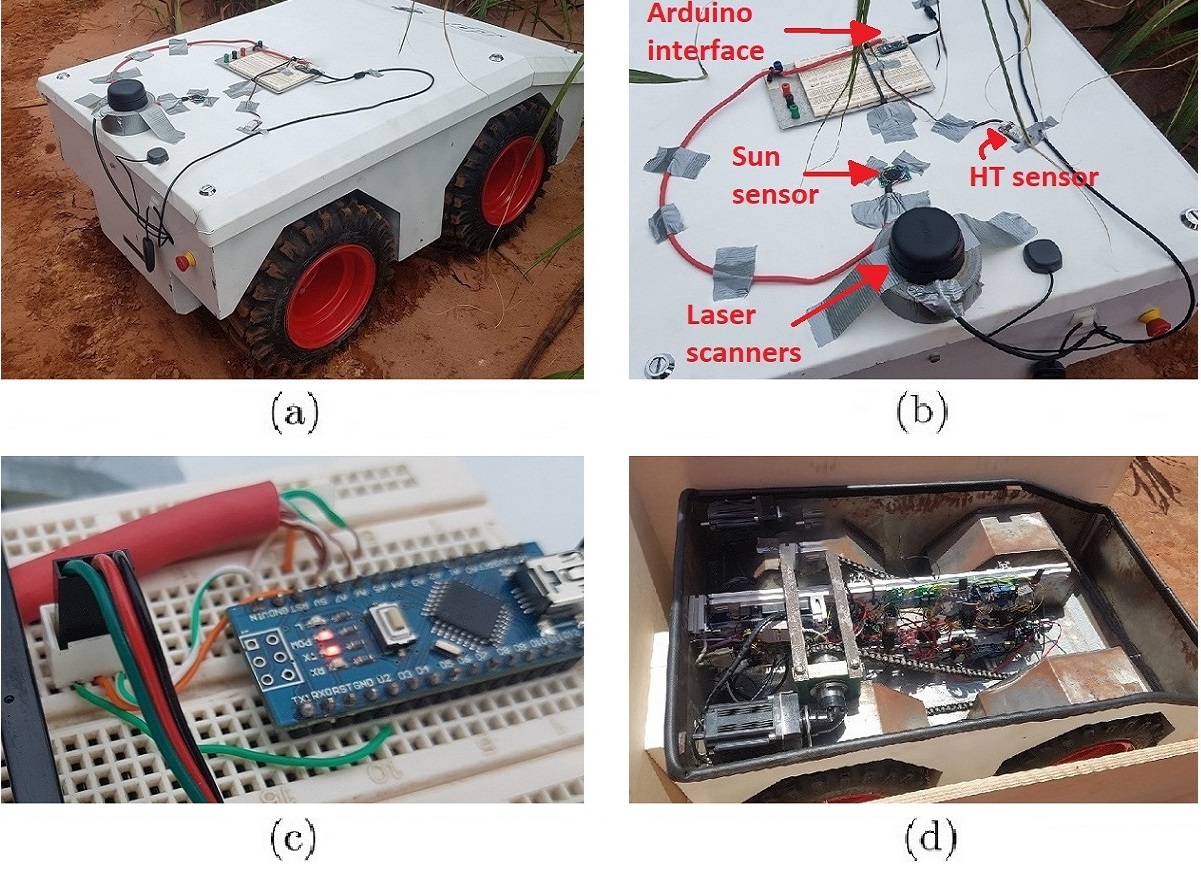}
	\caption{Preparation of robot for field test - (a) robot in the field; (b) sensors attached \emph{ad hoc} to the top lid; (c) interface for connecting solar sensor and HT sensor to Arduino; (d) electro-mechanical arrangement for the test. } 
	\label{FIG:FIELD1}
\end{figure}

The first field tests were performed at the UMOE    BioEnergy
sugarcane farm, at Presidente Prudente, in the state of São Paulo, Brazil. 
The main motivations for the farm visit and the robot   tests
were: evaluation of the mechanical concept and       proposed
dimensions, better examination and learning of            the
environment, and investigation of future modifications    and
new features,
as shown in
Fig.\,\ref{FIG:FIELD1}a.
%
A few sensors were initially employed in this field  test,
so that the understanding of the environment could         be
quantified in several data to be later analyzed and used   to
help the investigation of methods for the robot    navigation, as shown
in 
Figures~\ref{FIG:FIELD1}b and~\ref{FIG:FIELD1}c.
%
%

The tests were performed in different rounds, being all   the
sensor data of each round recorded in a different 
{\tt .bag}   file
in the ROS environment. 
For this test, different chains/sprocket configurations  were
tested, as shown in 
Fig.\,\ref{FIG:FIELD1}d.
%
%
\subsection{Mechanical Concept Evaluation}  
\begin{figure}[htpb] 
	\centering
	\includegraphics[angle=0,width=0.95\columnwidth]{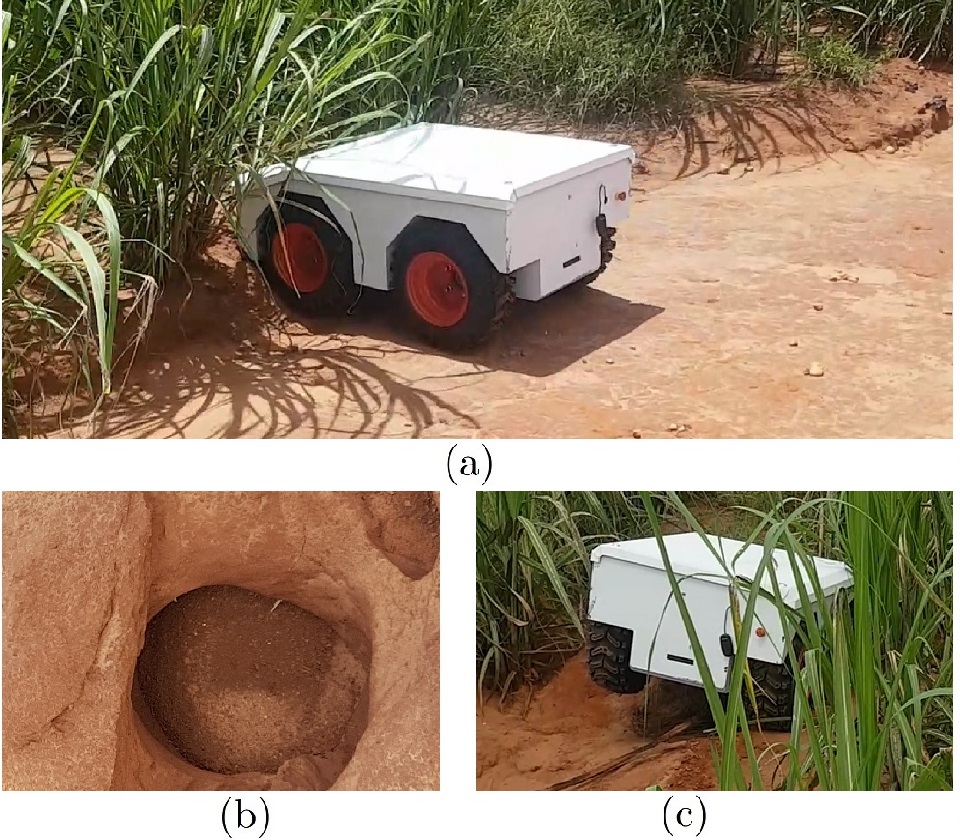}
	\caption{Field test in the UMOE BioEnergy farm, Presidente Prudente, Brazil - (a) robot entering the tunnel; (b) uneven terrain and big crevices; (c) robot passing over crevice, the structure is tilted and struggles to overcome the valley. } 
	\label{FIG:FIELD2}
\end{figure}

The robot presented a fairly 
satisfactory performance for remote navigation
over the sugarcane operational terrain, composed of sand  and
clay.
As expected, its behavior was rough
for operations in 
concrete and asphalt, because           the
skid-steering driving configuration and the selected    wheel
type require that the tire slides over the           terrain,
as shown in Fig.\,\ref{FIG:FIELD2}a.
The structure proved 
to be   robust
to the uneven terrain, and had a remarkable 
performance
for unexpected crevices on the                           soil,
as shown in 
Figures~\ref{FIG:FIELD2}b and \,\ref{FIG:FIELD2}c,
%
which have
been formed due to the accumulation of rain water in     that
period.  

\begin{figure}[htpb] 
	\centering
	\includegraphics[angle=0,width=0.95\columnwidth]{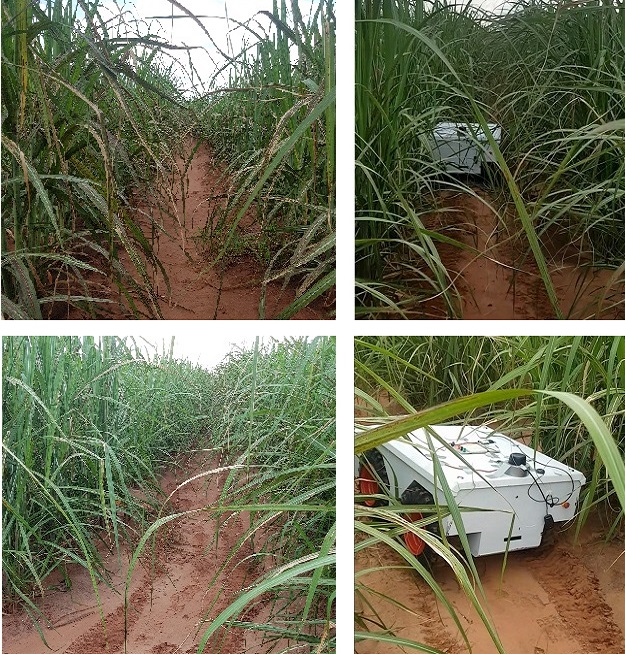}
	\vspace{-2mm}
	\caption{Field test in UMOE Bioenergy farm, Presidente Prudente SP, Brazil - Tunnels of plants. } 
	\label{FIG:FIELD3}
\end{figure}

The skid-steering arrangement showed to be very convenient to
achieve the desired motion and degrees-of-freedom:   forward,
backwards, rotation around own axis, and curves.          The
infiltration through the sugarcane tunnels was tested     for
plants with approximately 1, 2 and 3 meters              high (Fig.\,\ref{FIG:FIELD3}). For the first two cases, the   robot
was able to travel along the tunnel without any obstacle  for
its motion. In the 3 meters high case, were the plants    are 
tall enough to fall back towards the tunnel center and create
soft obstacles, there were some points in which the     robot
width seemed to be wide for this tightness and got the   risk
to collide with the plant stems. For all the cases, it    was
not observed the occurrence of plant leaves trapping       or
getting stuck to the structure or the wheels. 
\vspace{-0.35cm}
%
%
\subsection{Thermal Mapping} 
A FLIR\textsuperscript{\textregistered} One V2       infrared
camera in an smartphone was attached to the robot hood     in
order to capture thermal images and videos of the   sugarcane
tunnel environment. This was firstly proposed as a cheap  and
efficient solution for the robot navigation, which        was 
already proposed and implemented in many recent works, as by 
\cite{fehlman2009mobile}. 

\begin{figure}[!htpb]
	\centering
	\includegraphics[angle=0,width=0.95\columnwidth]{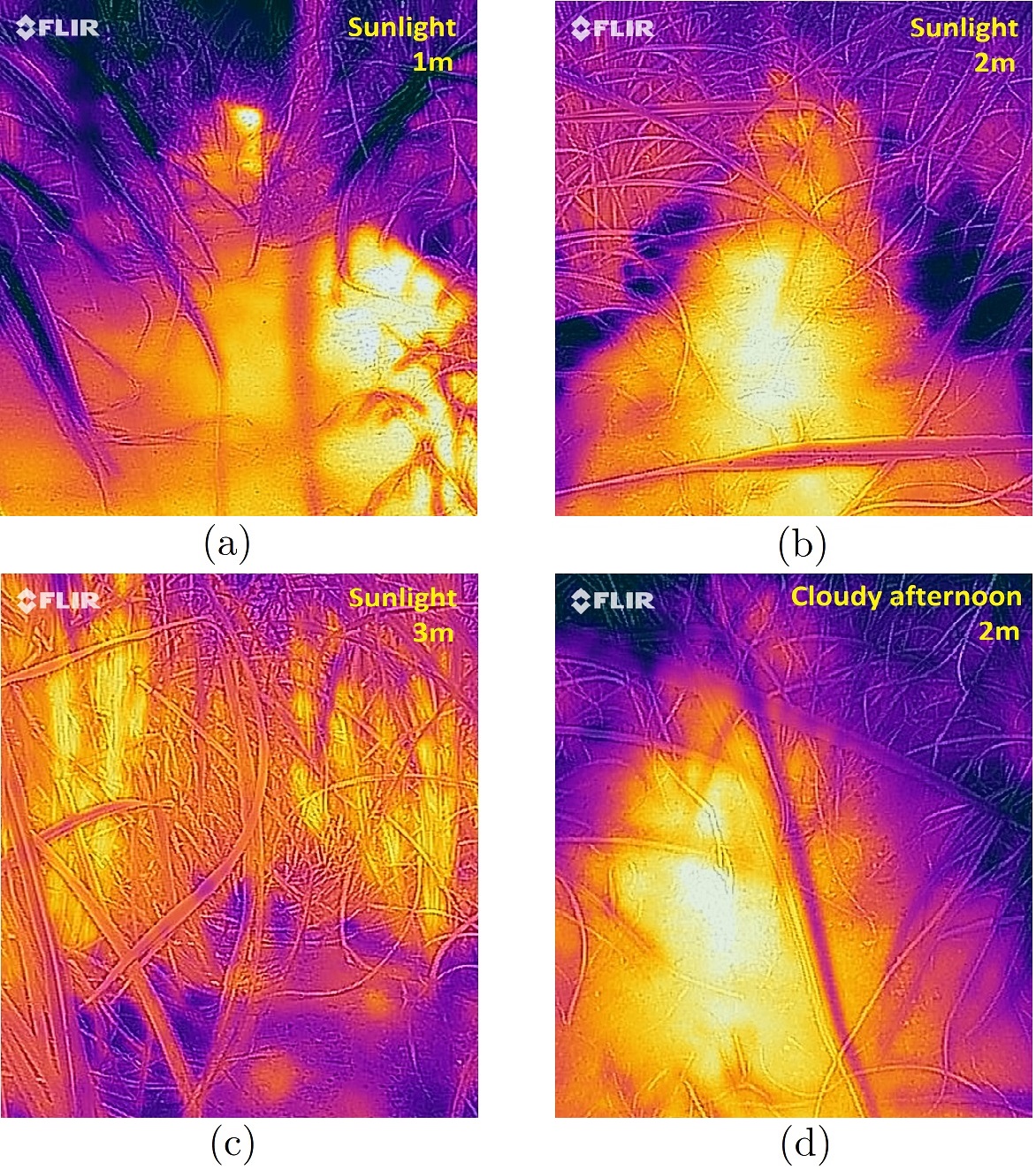}
	\caption{Thermal map - (a) sunlight, $1\,m$ plants; (b) sunlight, $2\,m$ plants; (c) sunlight, $3\,m$ plants; (d) late afternoon, cloudy, $2\,m$ plants. } 
	\label{FIG:THERMALMAP}
\end{figure}

The test was applied for rides through sugarcanes        with
approximately 1, 2 and 3 meters of height, and during     the
sunlight period - around 3 p.m. - and in the late   afternoon 
- around 6 p.m. and cloudy. The collected data were   greatly
satisfactory. The results for the sunlight             period  
(Fig.\,\ref{FIG:THERMALMAP}) show that the thermal     mapping
creates a path along the ground due to the        temperature
difference between the land and the plants. 
If compared to the image of a regular HD camera, the  thermal
path is much more distinguishable, and remains clear     even
when plants are trapping in front of the camera. For 1 and  2
meters tall plants, the ground is much warmer than        the
sugarcane, and a cold kerb is also created at the path border,
since the base of plantation is even colder. For 3     meters
tall plants, the sunlight does not reach the ground strongly. 
Therefore, the thermal map shows that the ground is    colder
than the plants and no kerb is visible. However, even with  a
smaller temperature difference, the created path is    strong
enough to be distinguished. 

\begin{figure}[!htpb]
	\centering
	\includegraphics[angle=0,width=1.0\columnwidth]{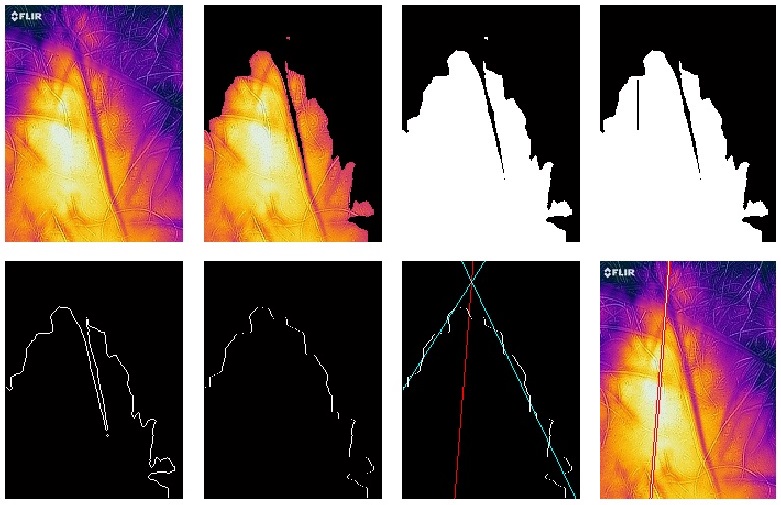}
	\caption{Example of thermal image processing performed with the collected data from the field test (the centerline is defined by linear regression of the margins of \emph{warm} and orange triangle output from a color threshold filter). } 
	\label{FIG:TERMO}
\end{figure}

The results for a cloudy                            afternoon,
as shown in 
Fig.\,\ref{FIG:THERMALMAP}d, 
%
%
and 2 meters                 tall
plants show that the heat pattern held strongly, so      that
clear paths were still visible. This was valid for all    the
three plant heights.  
Even though the thermal mapping were not also performed    in
rainy conditions, the collected results were          greatly
satisfactory, since navigation by using a small       thermal
camera would be a suitable low-cost solution, which is one of
the robot concept pillars. In addition to the taken   thermal
images and videos taken from the forward view, some media was
also collected with the camera sightly turned to the sides. 
The data can be used as an input for image processing and detection of the corridor centerline to be followed by the robot. Several techniques could be employed for that, such as simple and low-cost ones (e.g., color threshold), as shown in Fig.\,\ref{FIG:TERMO}, or complex ones (e.g., training artificial neural network for image classification).  

\subsection{Odometry and Laser Scanner} 
A 3D environment of the robot model and motion was created in
ROS by using the robot odometry. In this model, the     robot
position, orientation, linear and angular speeds           are 
calculated from the motors odometry. Typically, the data from
both encoders and a hall effect sensor are used, but        the 
current prototype still carries only the hall effect  sensors 
due to device delivery delays. The precision obtained by  the
hall sensor effect was considered satisfactory for        the
implementation of the first 3D model. The goal of the      3D 
environment is to make possible the simulation of the   robot 
motion, the visualization of sensor collected          data - 
depending on the used platform - and 
hence the investigation    of 
different techniques for the robot navigation and autonomy. 

\begin{figure}[htpb]
	\centering
	\includegraphics[angle=0,width=0.95\columnwidth]{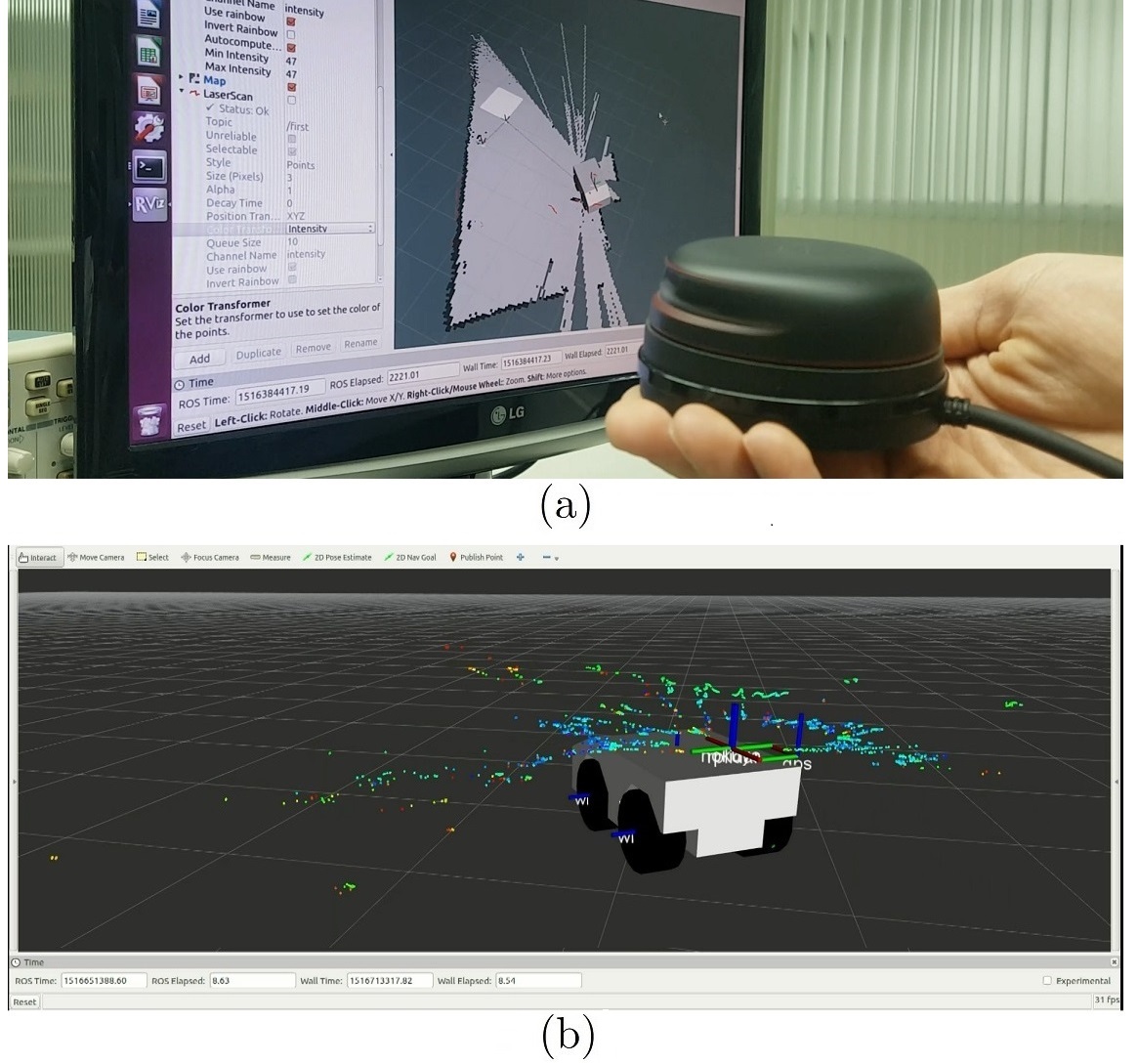}
	\caption{Laser data and 3D environment mapping - (a) LIDAR; (b) Hokuyo. } 
	\label{FIG:LASER1}
\end{figure}

\begin{figure}[htpb]
	\centering
	\includegraphics[angle=0,width=0.95\columnwidth]{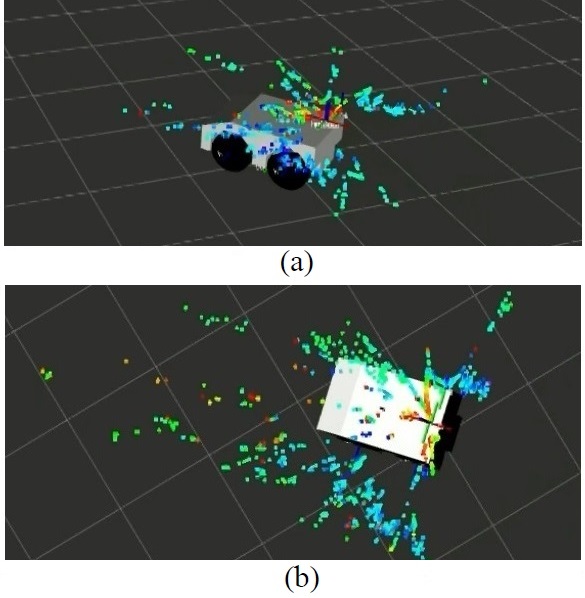}
	\caption{Corridor of points formed due to the existing corridor of plants (sugarcane tunnel) - (a) perspective view; (b) top view. } 
	\label{FIG:LASER2}
\end{figure}

For this prototype, the data of two laser scanners   attached
to the robot top lid - LIDAR\textsuperscript{\textregistered} 
A2 and Hokuyo\textsuperscript{\textregistered} -         were 
collected and viewed in the 3D environment together with  the 
robot model (Fig.\,\ref{FIG:LASER1}). 
Notice in Figure\,\ref{FIG:LASER2} that an         interesting
corridor of points was observed with the visualization of the
Hokuyo laser data, which is the scanning of the      existing
plant corridor deployed along the sugarcane tunnel.      Even 
though the precision of those laser scanners may not be   the
most recommended for this application - due to the       high
density of sugarcanes in such an environment - the   observed
data clearly motivates the use of those devices to aid in   a 
future robot mapping system.  

%
\subsection{Solar Sensor} 
\begin{figure}[htpb]
	\centering
	\includegraphics[angle=0,width=1.0\columnwidth]{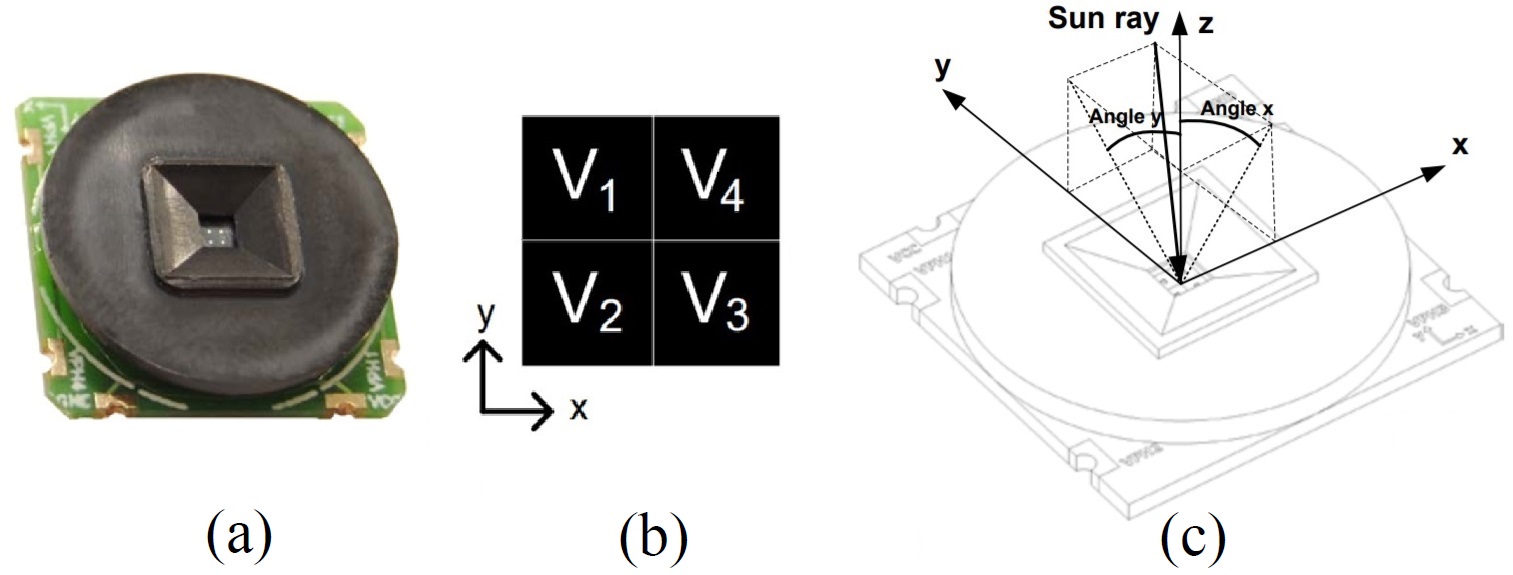}
	\caption{(a) Sensor NANO-ISSX; (b) photodetector cells; (c) sunlight projections along axes. 
	} 
	\label{FIG:SOLARSENSOR}
\end{figure}

One Solar Mems\textsuperscript{\textregistered}  NANO-ISSX-60
sun sensor was attached to the robot top lid in order      to
detect the direction of the                          sunlight 
(Fig.\,\ref{FIG:SOLARSENSOR}). The sensor power and    signals
were connected directly to the Arduino, and them communicated 
to the robot computer via CAN network. 
Solar 
sensors consist of a simple         embedded
technology that make possible the calculation of the sun  ray 
incident vector given its projection angles along  orthogonal 
axes. 
The sunlight is detected by a quadrant photodetector   device
aligned to the $x$- and $y$-axes, from where four     voltage 
levels denoted by $V_{1}$, $V_{2}$, $V_{3}$, $V_{4}$      are 
promptly read. 
The projection angles $\alpha_{x}$ and $\alpha_{y}$ are hence 
given by:
%
\begin{equation}
\alpha_{x}=\tan^{-1}{(C\,F_{x})} \,, \quad \quad 
\alpha_{y}=\tan^{-1}{(C\,F_{y})} \,, 
\end{equation}
where
%
%
%
\begin{eqnarray}
\label{eq2}
F_{x}&=&\dfrac{V_{1}+V_{2}-V_{3}-V_{4}}{V_{1}+V_{2}+V_{3}+V_{4}} \,, \\
&&  \nonumber \\
F_{y}&=&\dfrac{V_{2}+V_{3}-V_{1}-V_{4}}{V_{1}+V_{2}+V_{3}+V_{4}} \,,
\end{eqnarray}
%
%
%
%
%
and $C$ is a parameter that depends on the selected   sensor
model \cite{josequero}. 

The use of solar sensors in the first tests was     initially 
motivated by their low-cost, simplicity for installation, and
some important works that employed them for research     with
odometry \cite{lambert2011visual}, attitude           control
\cite{ortega2010miniaturized}, and                 navigation
\cite{volpe1999mars}. 
For this robot, the main motivation was to test the device in
the sugarcane environment and track the behaviour of      the
collected result in both sunny and cloudy weather  conditions 
and uneven terrain, which can cause a            considerable
disturbance in the expected results. 
One of the test goals is to determine the feasibility      of
using solar sensors to aid the robot initial heading      for
navigation. Another goal is related to a possible      future
improvement for this robot, which is the use of solar  panels
for the power supply. The solar sensor may help to    control
them to an efficient orientation configuration that can   get
the most out of the present sunlight.    
%
\subsection{Temperature and Humidity Sensor} %
A temperature and relative air humidity (HT) sensor -   model
RHT03 - was attached to the robot top lid, and   electrically
connected to the Arduino. HT sensors are strongly used in self-monitoring of embedded
electronics systems against
overheat and corrosion due to sea air. 
Their use was motivated for our environment,  since
it is subjected to high temperatures and humidity due to the
rain. Future implementations may be motivated by the use  two 
HT sensors: one within the chassis for the        electronics
self-monitoring, and another on the robot surface for       the
environment monitoring.     
%
\section{Concluding Remarks and Perspectives} 
%
%
The designed robot showed to be an efficient         low-cost
solution for deployment in gigantic sugarcane fields,    with
good motivation for operation in swarm-based approach. Since the conceptual test phase of the robot design was concluded and the
collected data was analyzed, some future developments  should 
be implemented. 
%
\subsection{Mechanical Modifications}
Although the mechanical concept presented a good performance,
the wheel fixation to the robot chassis is worryingly  stiff, 
while the chassis by itself is worryingly flexible.         
A proposed modification for the next robot version is      to
utilize steering axis damper along the wheel axes, and    a 
steel bar web crossing the chassis interior in order       to
enhance the structure stiffness. The selected wheel   bearing
model for the first prototype required the use of     several
{\it ad-hoc} mechanical adapters, which increased the   robot
weight, cost and complexity. 

For the next version, a new wheel bearing model will be used. 
The right-angle gearbox showed to be a good choice for the mechanical arrangement. Alternatively, we can use          a
straight-angle gearbox embedded in the wheel, which is      a 
more compact solution. A more suitable and           economic
power/motor selection should be performed for the        next 
version, because this was oversize for the first   prototype, 
given that the power requirement scenario was still unknown. 
The robot width should be reduced so that the robot can   fit
between the sugarcane tunnels in higher plant scenarios, such
as 3 meters tall tunnels. Finally, some original   design
features should be brought back, such as: aluminum    carcass
walls - in order to reduce the weight and power demand -  and
belts/pulleys in place of chains and sprockets -   in
order to reduce noise and backslash.     
%
\subsection{Improvement of Embedded Electronics} 
The embedded electronics system showed to be robust to vibrations and impacts. However, its main components     were
also oversized for the first prototype in terms of power  and
processing capacity. Planning the cost reduction for the next
robot version, the computer should be replaced by a      more
inexpensive solution, such as a Raspberry Pi 3. 
The VSS system will be composed of        a
customized printed circuit board (PCB) based on ATMEL     AVR
microcontrollers, which will replace the Arduino, the current
device power connections and many wiring. This will    better
organize the power distribution and enlarge            the
possibilities of communication, integration of new   sensors,
and programming. 
Originally, the idea of using solar photovoltaic (PV)  panels
for the power supply was being studied. However,    for
the current design, they became unrealistic, if       we 
consider the technology efficiency limitations and the  robot
dimensional design constraints. 

In future developments close to the available      commercial
version, the robot can have significantly weight    reduction
and more efficient power drivers. This can bring,    together
with possible and notable advances in Solar PV    technology,
the possibility of re-including this in the  robot
design and giving to it the expected power autonomy. 
%
\subsection{Software, Control and Navigation} 
The software architecture should go through           several
implementations in order to provide the expected features for
the next robot versions, such as: trajectory         control,
navigation, communication of computer with                the
VSS/self-monitoring, and autonomy. The trajectory control  is
the first step for the implementation of the robot  autonomy,
since it should be able to follow a predefined path along the
sugarcane tunnel. For the navigation through the    tunnels,
different  techniques based on the available sensors are  being
investigated, such as: following a corridor of         points
generation by the        laser scanner mapping, following   a
path defined by the thermal mapping, and using GPS/IMU     to
follow a path of known  points with predefined GPS geographic
coordinates, which are given by UMOE BioEnergy.           The
combination of those methods  together with robot     mapping
techniques may boost the definition of   an efficient     and
low-cost solution for the robot navigation. 
%
\subsection{New Features} 
\begin{figure}[htpb]
	\centering
	\includegraphics[angle=0,width=1.0\columnwidth]{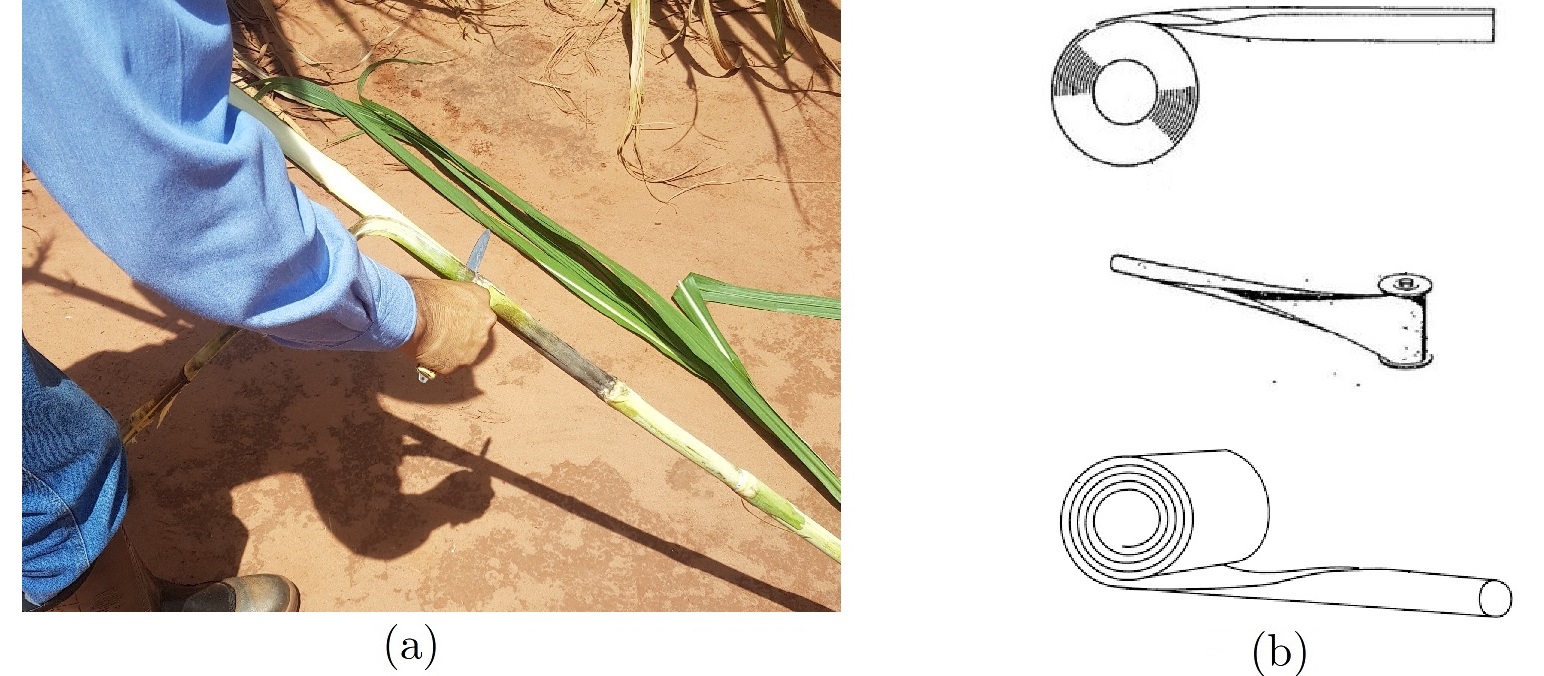}
	\caption{Sugarcane sampling in UMOE BioEnergy fields - (a) current procedure, the stem should be sliced or cut crosswise; (b) STEM, a compact solution for a lightweight sampling arm. 
	} 
	\label{FIG:SAMPLE}
\end{figure}

During the farm visit, some demand for new features were
investigated, being the collection of sugarcane samples   and
pest control the most remarkable. The sampling is a  frequent
procedure performed in the UMOE BioEnergy fields, which  aims
to detect pests in the sugarcane. 
This is a tedious procedure subjected to    extreme
weather, since the company staff may need to travel a    long
distance to the desired sampling point within the farm  area,
infiltrate the sugarcane tunnels and slice lengthwise      or
crosswise one piece of the sugarcane                     stem (Fig.\,\ref{FIG:SAMPLE}a).

For the next robot version, one possible solution for    this
problem is to attach a low-cost lightweight robotic arm    on
the robot chassis. Since this is a simple and low-payload task,
there is no need of a complex structure with          several
degrees-of-freedom. 
An interesting solution to be investigated is      a
serial manipulator formed by multiple Storable        Tubular
Extendible Members   (STEM\textsuperscript{\textregistered}, U.S. patents 3,144,215 and 3,434,674),
which are measure tapes that extend as     prismatic
joints and retracts inside their chambers in a    compact
solution, as shown
in Fig.\,\ref{FIG:SAMPLE}b.
%
Few studies about
this structure can be found in the literature, such as \cite{stem1}.
Finally, the pest control is a required feature to   overcome
the frequent problem of pest incidence.   Some
approaches for pest detection are being investigated,    such
as a drone for capturing aerial images with the    Normalized
Difference Vegetation Index (NDVI) applied. The collected
images of the same sugarcane field at UMOE BioEnergy farm
showed to be promising to detect the location of pest     and
send the robot directly to those points for a         precise
application of the pest control substance.    
%
\bibliography{cba}
\end{document}